\documentclass[a4paper, 10pt, conference]{IEEEtran}
\IEEEoverridecommandlockouts
\usepackage{cite}
\usepackage{amsmath,amssymb,amsfonts}
\usepackage{algorithmic}
\usepackage{graphicx}
\usepackage{textcomp}
\usepackage{xcolor}
\usepackage{ulem}
\def\BibTeX{{\rm B\kern-.05em{\sc i\kern-.025em b}\kern-.08em
    T\kern-.1667em\lower.7ex\hbox{E}\kern-.125emX}}
\begin{document}

\title{State Definition for Conflict Analysis with Four-valued Logic\\
}

\author{\IEEEauthorblockN{1\textsuperscript{st} Yukiko Kato}
\IEEEauthorblockA{ 
\textit{Lynx Research Llc}\\
Tokyo, Japan \\
ORCiD:0000-0003-0872-3847}

}

\maketitle

\begin{abstract}

We examined a four-valued logic method for state settings in conflict resolution models.
Decision-making models of conflict resolution, such as game theory and graph model for conflict resolution (GMCR), assume the description of a state to be the outcome of a combination of strategies or the consequence of option selection by the decision-makers.
However, for a framework to function as a decision-making system, unless a clear definition of the task of placing information out of an infinite world exists, logical consistency cannot be ensured, and thus, the function may be incomputable.
The introduction of paraconsistent four-valued logic can prevent incorrect state setting and analysis with insufficient information and provide logical validity to analytical methods that vary the analysis resolution depending on the degree of coarseness of the available information.
This study proposes a GMCR stability analysis with state configuration based on Belnap's four-valued logic.

\end{abstract}


*\textit{This work has been submitted to the IEEE for possible publication. Copyright may be transferred without notice, after which this version may no longer be accessible.}

\section{Introduction}
In the decision analysis framework, a common method involves setting the status quo and obtaining a ranking of the preferred states for a decision-maker (DM) by repeating the pairwise comparison of states that differ from the status quo and are more preferable for the DM. When the preferences of multiple DMs' conflict, some suggestions for solutions to the conflict can be obtained by synthesizing these preferences and estimating the states in which the entire situation is likely to settle.
That is, the methodology we use to recognize and describe the initial state is the starting point for all subsequent classifications, analyses, and evaluations.
Meanwhile, most of the discussions in decision-making systems focus on how to derive and expand the concepts related to preferences and stability; however, an in-depth discussion regarding the state perception and definition does not exist.
This study aims to scrutinize the mechanism for the definition of states in conflict resolution models, especially the graph model for conflict resolution (GMCR), discuss the problems, introduce findings based on the Kripke structure, and propose a new method of setting the state.

The following section clarifies the basic concepts and background of the methods used in this study for decision and conflict resolution analysis by overviewing the literature; essential definitions and frameworks are provided.
Section \ref{new} explains the concepts and frameworks that are proposed in this paper. Next, in Section \ref{sec:4}, the new framework presented in Section \ref{new} is applied to the conflict cases for further observation and analysis.
In the Conclusion section \ref{concl.}, the future research aspects based on the results of this study are described.

\section{Underlying Concepts and Methods}
GMCR, the basic framework of this paper, was derived from non-cooperative game theory , and was developed by D.M. Kilgour, K.W. Hipel, and L. Fang\cite{interaction},\cite{gmcr}.
Extended concepts of GMCR include coalition analysis \cite{coalition-1},\cite{coalition-2},\cite{coalition-3}, attitude-based analysis \cite{attitude-1},\cite{attitude-2}, and preference strength\cite{strength}; furthermore, research has been actively conducted using fuzzy\cite{fuzzy-1}, \cite{fuzzy-2}, \cite{fuzzy-3} and grey methods\cite{grey} for cases in which the DM preferences are uncertain.
Other analysis methods with coarser information have been proposed, such as ``permissible ranges'' for preferences and analyzing them with binary information \cite{kato-1}, \cite{kato-2}.

GMCR adopts the concept that DMs can transition to a state that is preferable to them within a choice of options that they can control. In game theory, the focus of the discussion is on the DM's ``strategy,'' but in GMCR, the focus is on the ``state'' and ``state transition'' as a consequence of option selection. The framework has more flexibility and broader descriptive capability owing to the unique concept of state transitions, which is not addressed in game theory. 

In decision-making, we generally deal with the following essential elements: a set of ``options,'' ``consequences,'' ``maps from the option selections to the consequences,'' and a ``preference structure for consequences.''
An option describes a situation in which decision-makers can make their own choice.
 In a conflict, the elements of a situation to be decided by other decision-makers are not options; similarly, elements that the decision-maker has not yet identified but could affect his situation, such as climate or global economic trends, are also not options.
 In GMCR, a state is a description of the consequence of the DMs' selection of options.
 The following subsections provide an overview of the basic framework of the GMCR.

\subsection{Graph Model for Conflict Resolution}
\subsubsection{Framework }
\label{Framework}
GMCR is a framework that can be described with four tuples\cite{interaction},\cite{gmcr}:
\begin{equation}
(N,S,(A_i)_{i\in N},(\succsim_i)_{i\in N}).
\end{equation}
$N$ is the set of all DMs involved in the conflict, where $|N| \geq 2 $. $S$ is the set of all possible states in the conflict, where $|S| \geq 2$. If $i \in N$, then the pairs $(S, A_i)$ constitute a graph of DM $i$, where $S$ is the set of all vertices and $A_i \subset S\times S $ is the set of all arcs. We also assume that $(S,A_i)$ has no loops, i.e., $(s,s) \in A$ for each $s \in S$.
$\succsim_i$ is the DM $i$'s preference for possible states, satisfying reflexivity (for any $s \in S, s \succsim_i s$), completeness (for any $s, s^{\prime} \in S, s \succsim_i s^{\prime}$ or $s^{\prime} \succsim_i s$), and transitivity (for any $s, s^{\prime},s'' \in S$, if $s \succsim_i s^{\prime}$ and $s^{\prime} \succsim_i s ''$, then $s \succsim_i s ''$).
A rational DM desires the situation to change to a more favorable state and attempts to move to the preferred states by repeating unilateral moves: transitions over which the DM has control.

As a consequence of the DMs’ moves in GMCR, standard stability concepts, including Nash stability (Nash)\cite{nash1951},\cite{nash1950}, general metarationality (GMR) \cite{howard}, symmetric metarationality (SMR)\cite{howard}, and sequential stability (SEQ) \cite{fraser-1},\cite{fraser-2}, exist; some underlying notions must be provided.
For $i \in N$ and $s \in S$, we define DM $i$’s reachable list from the state $s$ as the set $\{s^{\prime} \in S\mid (s, s^{\prime}) \in  A_i\}$, denoted by $R_i (s)$.  $R_i(s)$ is the set of all states in the conflict that DM $i$ can move from $s$ to $s^{\prime}$ in one step.
A unilateral improvement of DM $i$ from state $s$ is defined as an element of the reachable list of
DM $i$ from $s$ (i.e., $s^{\prime} \in R_i (s)$), where DM $i$ strictly prefers state $s^{\prime}$ ($ s^{\prime} \succ_i s$).
Therefore, the set of unilateral improvement lists of DM $i$ from state $s$ is the set $\{s^{\prime} \in R_i(s)| s^{\prime} \succ_i s\}$, which is known as DM $i$’s unilateral improvement list from state $s$ and denote it by $R^+_i (s)$. 
$\phi^{\simeq}_i(s)$ denotes the set of all states that DM $i$ equally or less prefers to state $s$; $\{s^{\prime} \in S\mid s \succsim_is^{\prime} \}$.

\subsubsection{Stability Concepts}
The standard stability concepts in GMCR: Nash, GMR, SMR, and SEQ can be defined based on the four tuples and the reachable lists noted in  \ref{Framework}.\\
\textbf{NASH}: For $i \in N$ and state $s \in S$, $s$ is a Nash state for DM $i$ if and only if $R^+_i(s)=\emptyset $. The set of all Nash states for DM $i$ is denoted $S_i^{Nash}$.
A Nash state of DM $i$ is considered stable for DM $i$ as DM $i$ cannot unilaterally improve from that state.\\  
\textbf{GMR}:
$s$ is the GMR state for DM $i$ if and only if $s^{\prime}\in R_i^+(s)$,  $R_{N\backslash i}(s^{\prime})\cap \phi^{\backsimeq}_i(s) \neq \emptyset$. DM $i$’s GMR state is stable for DM $i$; each DM $i$ is deterred from unilaterally improving from that state as the countermove of other DMs leaves DM $i$ in a state equal to or less favorable than its original state. The set of all GMR states for DM $i$ is denoted as $S_i^{GMR}$.\\
\textbf{SMR}:
$s$ is an SMR state for DM $i$, if and only if for all $s^{\prime} \in R^+_i (s)$, there exists $s'' \in R_{N\backslash {i}}(s^{\prime}) \cap \phi_i^{\backsimeq}(s) \neq \emptyset$ such that $s^{\prime\prime\prime} \in \phi^{\backsimeq}_i(s)$ for all $s^{\prime\prime\prime} \in R_i(s'')$, i.e., $R_i(s'')\subseteq \phi^{\backsimeq}_i(s)$.
The set of all SMR states for DM $i$ is denoted $S^{SMR}_{i}$. In DM $i$'s SMR states, the countermove of the other DMs on the unilateral improvement of DM $i$ will result in DM $i$ achieving a state that is no better than the original state even if DM $i$ unilaterally moves from that state.
Consequently, the original state can be considered stable for DM $i$ because the first unilateral improvement is not beneficial for DM $i$.\\
\textbf{SEQ}:
$s$ is an SEQ state for DM $i$ if and only if for all $s^{\prime} \in R^+_i(s), R^+_{N\backslash {i}}(s^{\prime}) \cap \phi^{\backsimeq}_i(s)= \emptyset$. 
The set of all SEQ states for DM $i$ is denoted as $S^{SEQ}_{i}$.
In the SEQ state of DM $i$, the credible countermove of other DMs on DM $i$’s unilateral improvement achieves a state that is equally favorable or less favorable than the original state to DM $i$. Therefore, the original state can be regarded as stable for DM $i$ as DM $i$ never becomes better than the original state.\\
For each of these stability concepts, $s \in S$ is considered to be in equilibrium if and only if $s$ is stable for all DMs.

Furthermore, CNash, CGMR, CSMR, and CSEQ have also been proposed as coalition stability concepts when the DM believes that a transition to a better state can be achieved by repeating moves in cooperation with other DMs.

\subsubsection{State}
In GMCR, the alternatives of actions that the DM can control are called options. The set of all options in a conflict is $O=\cup_{i\in N} O_i,$ where $i$ denotes which DM controls the options. We let $O_i$ represent the set of options of DM $i$ for $i \in N$ with $o_{ij} \in O_i$.
An option selection for DM $i$ is a mapping $\lambda: O_i\rightarrow \{0,1\}$, such that for $j=1,2,\dots,h_i$, where $o_{ij}$ is DM $i$'s $j$th option.
\begin{equation}
\lambda(o_{ij})=\begin{cases}
        {1 \: {{{\rm {if}\:\: DM}\,i\, {\rm {selects \, option}}}}\, o_{ij}},\\
        {0 \: \rm {otherwise}}.
    \end{cases}
\end{equation}
Let $O=\bigcup_{i \in N}O_i$ be the set of all options for $o_{ij} \in O_i, i=1,2, \dots, n.$ A state is a mapping $\kappa:O \rightarrow \{0,1\}$, such that for $i=1,2,\dots,n$.
\begin{equation}
\kappa(o_{ij})=\begin{cases}
        {1 \: {{{\rm {if}\:\: DM}\,i\, {\rm {selects \, option}}}}\, o_{ij}},\\
        {0 \: \rm {otherwise}}.
    \end{cases}
\end{equation}
Let $|O|$ represent the total number of options available to the DMs. A state can be treated as a $|O|$-dimensional column vector with 0 or 1 as the element. $\lambda^s(O_i)$ denotes DM $i$’s option selection corresponding to state $s$ for $i= 1,2,\dots,n$ and is a $|O_i|$-dimensional column vector whose elements are as follows:
\begin{equation}
    \lambda^s(o_{ij})=\begin{cases}
        {1 \: {{{\rm {if}\:\: DM}\,i\, {\rm {selects \, option}}}}\, o_{ij} \:\:{\rm {in \:\:state}}\:\:s },\\
        {0 \: \rm {otherwise}}.
        \end{cases}
\end{equation}
Prisoners' dilemma can be illustrated as in Table \ref{tab:PD1} , where ``Y'' indicates that an option is selected by the DM, whereas ``N'' implies that the option is not selected.

\begin{table}[t]
\caption{Prisoners' Dilemma}
\label{tab:PD1}
\centering
\small
\begin{tabular}{|c|c|c|}
\hline
  {}    &  {\textbf {DM1}}&  {\textbf {DM2}} \\
\hline                           	
 Option & Not Confess & Not Confess \\
\hline
$s_1$ & Y & Y\\
$s_2$ & Y & N\\
$s_3$ & Y & Y\\
$s_4$ & Y & N\\
\hline
\end{tabular}
\end{table}

The option form or binary form as shown in Table \ref{tab:PD1}, which expresses the conflicts based on the selection of options, was initially proposed by Howard \cite{howard}. Furthermore, the option form is considered one of the methods of describing a state in the framework of GMCR, which is the underlying conceptual foundation \cite{interaction}.

\subsection{Configuration of States and Information Content }
In GMCR, the preference order for states is derived either by direct evaluation of each state or by a method that satisfies the first-order predicate logic that assumes unconditional (negation, conjunction, separation) and conditional (iff) combinators to specify the option-prioritization method order.
In addition to fuzzy and grey, unknown \cite{unknown-1}, \cite{unknown-2},  and probabilistic methods \cite{probabilistic} have been studied when the preference information is ambiguous or unknown; none of the above methods provide conceptual clarifications regarding the basis for the states, but they focus on the viewpoint of the uncertainty of preferences.
Nevertheless, when analyzing real-world problems, information may be insufficient at the outset of the analysis to define states with sufficient logical consistency to withstand analysis, all at the same level of granularity.
Therefore, we consider a framework that allows for ambiguity and uncertainty in setting the starting point of the analysis. With such framework, we need not rewrite the original setting, but we can refine the information granularity when the situation develops and more information is available.

\section{State Definition by Four-valued Logic}
\label{new}
A state in GMCR is represented as a combination of option selection by DMs expressed as a binary value of whether DM $i$ has selected a particular option.
To define states in a manner more similar to real-world state recognition, we consider states as information granules and an analysis system that can simultaneously handle the states expressed in different information granularities.
Regarding the state definition, i.e., ``framing'' of real-world problems, methods based on four-valued logic have been studied by Nakayama et al.\cite{nakayama} mainly from the perspective of computability.
The four-valued logic uses the concept of granularity reasoning by introducing variable precision rough set models (VPRS) by Ziarko \cite{VPRS} to the possible world concept in Kripke and describes it in terms of four values, was developed by Belnap \cite{belnap}.
\subsection{Belnap's Four-valued Model}
Belnap's four-valued set, represented by \textsf{B4}, comprises the elements of truth value: \{\textbf{T}, \textbf{F}, \textbf{B(oth)}, \textbf{N(one)}\}, where \textbf{N} indicates incompleteness, and \textbf{B} indicates contradiction. The current state of an atomic proposition in the computer can be described by four values: true, false, true and false, and neither true nor false.
The values for complex sentences were obtained by Balnap by considering the monotonicity, which yields Table \ref{tab:4table}.

\begin{table}[t]
\caption{\textbf{4}-Truth Table }
\label{tab:4table}
\centering
\begin{tabular}{|c|c|c|c|c|}
\hline
{} & \textbf{N}& \textbf{F}& \textbf{T} & \textbf{B} \\
\hline
$\neg$ & \textbf{N}& \textbf{T}& \textbf{F} & \textbf{B} \\
\hline\hline
\end{tabular}

\centering
\begin{tabular}{|c|c|c|c|c|}
\hline
 $\wedge$ & \textbf{N}& \textbf{F}& \textbf{T} & \textbf{B} \\
\hline
\textbf{N} & \textbf{N}& \textbf{F}& \textbf{N} & \textbf{F} \\
\hline
\textbf{F} & \textbf{F}& \textbf{F}& \textbf{F} & \textbf{F} \\
\hline
\textbf{T} & \textbf{N}& \textbf{F}& \textbf{T} & \textbf{B} \\
\hline
\textbf{B} & \textbf{F}& \textbf{F}& \textbf{B} & \textbf{B} \\
\hline\hline
\end{tabular}

\centering
\begin{tabular}{|c|c|c|c|c|}
\hline
 $\vee$ & \textbf{N}& \textbf{F}& \textbf{T} & \textbf{B} \\
\hline
\textbf{N} & \textbf{N}& \textbf{N}& \textbf{T} & \textbf{T} \\
\hline
\textbf{F} & \textbf{N}& \textbf{F}& \textbf{T} & \textbf{B} \\
\hline
\textbf{T} & \textbf{T}& \textbf{T}& \textbf{T} & \textbf{T} \\
\hline
\textbf{B} & \textbf{T}& \textbf{B}& \textbf{T} & \textbf{B} \\
\hline
\end{tabular}

\end{table}

\subsection{State Definition by Four-Valued Logic}
Next, we define the states in Prisoner's dilemma using \textsf{B4} by viewing the option proposition ``DM $i$ does not confess'' as an atomic statement. Table \ref{tab:PD states} represents sixteen combinations of propositions generated by \textsf{B4} .

\begin{table}[t]
\caption{Prisoner's Dilemma -States by \textsf{B4}}
\label{tab:PD states}
\centering
\footnotesize
\begin{tabular}{c|c|c}
\hline
{}  & \textbf{DM1} &\textbf{DM2} \\
$p$ & Not Confess &  Not Confess \\
\hline\hline
$1$ & \textbf{T} & \textbf{T}  \\
$2$ & \textbf{T} &  \textbf{F} \\
$3$ & \textbf{T} & \textbf{N} \\
$4$ & \textbf{T} & \textbf{B}  \\
$5$ & \textbf{F} & \textbf{T}  \\
$6$ & \textbf{F}  & \textbf{F} \\
$7$ & \textbf{F}  & \textbf{N}  \\
$8$ & \textbf{F}  & \textbf{B}  \\
$9$ & \textbf{N}  &  \textbf{T}  \\
${10}$ & \textbf{N}  & \textbf{F} \\
${11}$ & \textbf{N}  & \textbf{N}  \\
${12}$ & \textbf{N}  &  \textbf{B}  \\
${13}$ & \textbf{B} &   \textbf{T}  \\
${14}$ & \textbf{B} &  \textbf{F}  \\
${15}$ & \textbf{B} & \textbf{N}  \\
${16}$ & \textbf{B}  & \textbf{B} \\
\hline
\end{tabular}
\end{table} 
    
 For example, suppose there is a rumor that a pardon may be issued soon, and the DM knows that only cases that have been closed after interrogation are eligible for the amnesty. In such a case, we can understand that the DM may choose \textbf{B}, which allows making an option depending on the status of the execution of the pardon.
 \textbf{N} could be viewed as a DM's situation in which the DM is planning to escape from prison without cooperating with the interrogation.
In the general GMCR method of analysis, states are set on the assumption that they can only be established by options that are definitely controllable by the DM; hence, no other possibilities are considered. As a way to describe other situations. Regarding this,
the concept of ``new reachability by external factors'' as a method to describe states involving non-optional elements that cannot be controlled by DMs was proposed by Kato \cite{kato-2}, \cite{kato-3}.
Meanwhile, it can be said that the state description by \textsf{B4} solves the problem of the potential state generation not by extending state transitions but by extending the state definition itself.
Unlike state generation with binary information, to set the preference order for each state in \textsf{B4} ,  DMs' policy for \textbf{B} and  \textbf{N} at the state setting or framing \cite{frame} would need to be established.
The preference of the DM may be based on the objective that he prioritizes terminating the interrogation in a shorter time, avoiding \textbf{B} or \textbf{N} regardless of the outcome of obtaining a pardon.
Alternatively, some DMs may not find the rumors of amnesty enforcement credible and thus would prefer option \textbf{B} or \textbf{N}.

\subsection{GMCR Stability Analysis by \textsf{B4} }
In order to perform GMCR stability analysis, reachability and preferences for each DM must be determined.
\subsubsection{Reachability}
Based on Anderson and Belnap's entailment \cite{entailment}: $\textbf{F}\preceq \textbf{B}$, $\textbf{F}\preceq \textbf{N}$, $\textbf{B}\preceq \textbf{T}$, $\textbf{N}\preceq \textbf{T}$, the reachability in GMCR can be defined.
Thus, we consider that DM1 can make a unilateral move when DM2’s proposition of the initial state and the destination state is any of the following pairs: (\textbf{B},\textbf{B}), (\textbf{B},\textbf{F}), (\textbf{N},\textbf{N}), (\textbf{N},\textbf{F}), (\textbf{T},\textbf{T}), (\textbf{T},\textbf{B}), (\textbf{T},\textbf{N}).

\begin{figure}[b]
 \centering
  \includegraphics[scale=0.5]{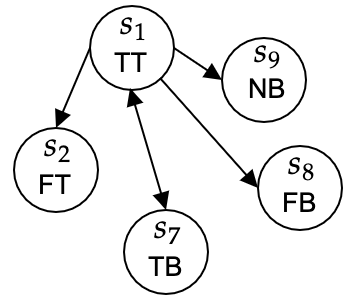}
\caption{DM1's Graph from $s_1$}
\label{fig:reachability}
\end{figure}

The Figure \ref{fig:reachability} illustrates DM1's reachability from $s_1$.

\subsubsection{Preference}
In addition to the usual preference information, each DM's attitude toward situations involving \textbf{B} and \textbf{N} must be assessed.

The results of the preference and stability analysis for the two DMs are listed in Table \ref{tab:PD_4}, assuming that DM1 has a preference to finish the interrogation as soon as possible in anticipation of pardon enforcement. In contrast, DM2 has no information on the pardon and thus has a preference not to make as clear a statement as much as possible and to be uncooperative during the interrogation.
The orders of preference for the two DMs were set as follows:
$s_2\succ_1 s_1\succ_1 s_4 \succ_1 s_3 \succ_1 s_8 \succ_1 s_6\succ_1 s_7\succ_1 s_5\succ_1 s_9\succ_1$, $s_7\succ_2 s_8\succ_2 s_9 \succ_2 s_5 \succ_2 s_6 \succ_2 s_3\succ_2 s_1\succ_2 s_4\succ_2 s_2\succ_2$.

\subsubsection{Stability Analysis}
For simplicity of the stability analysis, we reduced the number of states is reduced from 16 to 9.

\begin{table*}[t]
\footnotesize
\caption{Prisoner's Dilemma in \textsf{B4}  - Stability Analysis }
\label{tab:PD_4}
\centering

\begin{tabular}{lccccccccc}
\hline 
{} & $s_1$ & $s_2$ & $s_3$ & $s_4$ & $s_5$ & $s_6$ & $s_7$ & $s_8$ & $s_9$ \\
\hline
DM1 & \textbf{T}  & \textbf{F} & \textbf{T} & \textbf{F} & \textbf{T} & \textbf{F}& \textbf{T} & \textbf{F} & \textbf{N}  \\

$R_1$ & 2,5,6,7,8,9 & 1,5,6,7,8,9 & {} & {} & 3,4,6 & 3,4,5 & 3,4,8,9 & 3,4,7,9 & 3,4,7,8\\
\hline
DM2 & \textbf{T}  & \textbf{T} & \textbf{F} & \textbf{F} & \textbf{N} & \textbf{N}& \textbf{B} & \textbf{B} & \textbf{B}  \\
$R_2$ & 3,5,7 & {} & 1,5,7 & {} & 1,3,7 & {} & 1,3,5 & {} & {}\\
\hline \hline
Nash & {}   & \textit{4*} & {} & \checkmark & {} & {}& {} & {}  & {}  \\
\hline

GMR &  \textit{2*}  & \textit{4*} & \textit{4*} & \checkmark  & {} & {}& {} & {}  & {}  \\
\hline

SMR &  \textit{2*}  & \textit{4*} & \textit{4*} & \checkmark  & {} & {}& {} & {}  & {}  \\
\hline

SEQ & \textit{2*} & \textit{4*} & \textit{4*} & \checkmark  & {} & {}& {} & {}  & {} \\
\hline

CNash & {}  & \textit{4*} & {} & \textit{4*} & {} & {}& {} & {}  & {}  \\
\hline
CGMR & \textit{2*}  & \textit{4*} & \textit{4*} & \textit{4*} & {} & {}& {} & {}  & {}  \\
\hline
CSMR & \textit{2*}  & \textit{4*} & \textit{4*} & \textit{4*} & {} & {}& {} & {}  & {}  \\
\hline
CSEQ & \textit{2*}  & \textit{4*} & \textit{4*} & \textit{4*} & {} & {}& {} & {}  & {}  \\
\hline
Pareto & \checkmark & \checkmark & \checkmark & {} & {} & {} & \checkmark  & \checkmark  & {} \\
\hline
\end{tabular}
\end{table*} 

In the Table \ref{tab:PD_4}, a checkmark indicates that equilibrium is established both in the standard form and in \textsf{B4}, \textit{4*} represents that equilibrium is shown only in \textsf{B4}. Finally, \textit{2*} indicates that equilibrium is found only in the two-valued standard form.
$R_1$ represents the destination states to which DM1 can transition from the state, and $R_2$ represents the destinations where DM2 can transition.

The analysis shows that the original dilemma in $s_4$ still holds in the new conflict setting with states represented in \textsf{B4}, whereas other equilibria are newly established in $s_2$ and $s_3$.
The results of the GMCR stability analysis using states according to \textsf{B4} are as follows: $S^{Nash}=\{s_2, s_4\}$, $S^{GMR}=\{s_2, s_3, s_4\}$, $S^{SMR}=\{s_2, s_3, s_4\}$, $S^{CNash}=\{s_2, s_4\}$ $S^{CGMR}=\{s_2, s_3, s_4\}$, $S^{CSEQ}=\{s_2, s_3, s_4\}$.
A strong equilibrium established in $s_2$, including a Nash equilibrium, should be noted.

\section{Reinterpretation of Conflicts in Four-Valued Logic}
\label{sec:4}
In the previous sections, we have presented the basic concepts of GMCR, the new concept of states definition by introducing the four-valued logic, and examples of its implementation.
This section provides a new perspective on Elmira conflict, a representative application analysis of GMCR.

\subsection{State 9 in Elmira Conflict}
The Elmira conflict is an environmental contamination dispute in Ontario, Canada; numerous studies have been already conducted on this conflict using the GMCR. Three DMs are involved in the conflict: the Ministry of Environment ($\textbf{M}$), Uniroyal ($\textbf{U}$), and the local government ($\textbf{L}$). $\textbf{M}$ discovered contamination and issued a control order to $\textbf{U}$ that included decontamination operation by $\textbf{U}$. They want to exercise their authority efficiently. $\textbf{U}$ owns questionable
chemical plants and intends to exercise its right to object, aiming to lift or relax the control order. $\textbf{L}$ represents the diverse interest groups and intends to protect the residents and the local industrial base.

\begin{figure}[t]
      \centering
 \includegraphics[scale=0.45]{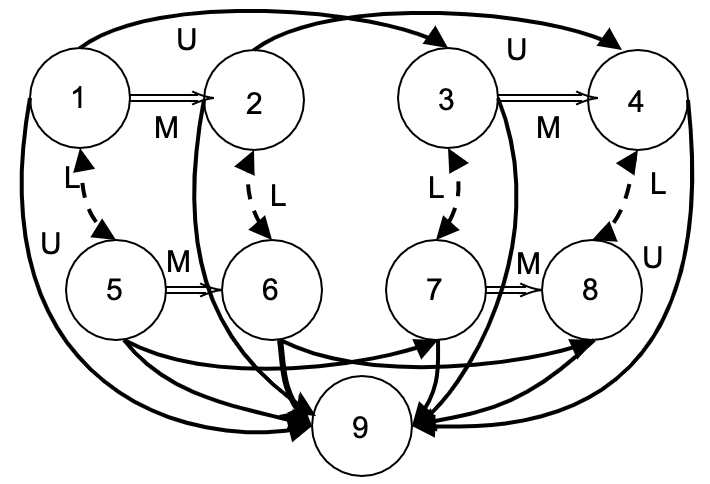}
  \caption{Graph Model of Elmira conflict}
      \label{fig:Elmira}  
 \end{figure}

The options for each DM are as follows. $\textbf{M}$: can
irreversibly modify the control order; $\textbf{U}$: may continue to delay the objection process, irreversibly accept the control order, or abandon the chemical plants irreversibly;
or irreversibly modify the control order; $\textbf{U}$: may continue to delay the objection process, irreversibly accept the control order, or abandon the chemical plants irreversibly;
$\textbf{L}$: may argue for the application of the original control order. Table \ref{tab:Elmira options} summarizes all the feasible states based on the DM's options, and Figure \ref{fig:Elmira} represents the Graph of the conflict. The numbers in the graph indicate the state numbers.

\begin{table}[t]
\caption{Elmira Conflict - Options and States}
\label{tab:Elmira options}
\centering
\small
\begin{tabular}{clccccccccc}
\hline
{} & {} & $s_1$ & $s_2$ & $s_3$ & $s_4$ & $s_5$ & $s_6$ & $s_7$ & $s_8$ & $s_9$ \\
\hline
M & Modify & N & Y & N & Y & N & Y & N & Y & - \\
\hline
U & Delay & Y & Y & N & N & Y & Y & N & N & - \\
    & Accept & N & N & Y & Y & N & N & Y & Y & -\\
    & Abandon & N & N & N & N & N & N & N & N & Y\\
 \hline
L & Insist & N & N & N & N & Y & Y & Y & Y & - \\   
\hline
\end{tabular}
\end{table} 

\textbf{U}'s second option where \textbf{U} abandons the plant,  which is the cause of the pollution, has the extreme significance in this case. $s_9$ in the graph reveals that \textbf{U} uses this option as a powerful bargaining card.
In $s_9$, no other DM option selections are provided. In other words, $s_9$ can be understood as the state in which \textbf{U} abandons the plant regardless of the other DM's option choices.
In other words, the structure of state 9 in  Elmira conflict can be interpreted as a state in which only U's ``abandon'' option value is \textbf{T}, while all other options are \textbf{B}. 
Table \ref{tab:Elmira_4} shows the 9 states in \textsf{B4}.

\begin{table}[t]
\caption{Elmira Conflict - States in \textsf{B4}}
\label{tab:Elmira_4}
\centering
\small
\begin{tabular}{clccccccccc}
\hline
{} & {} & $s_1$ & $s_2$ & $s_3$ & $s_4$ & $s_5$ & $s_6$ & $s_7$ & $s_8$ & $s_9$ \\
\hline
M & Modify & \textbf{F} & \textbf{T}  & \textbf{F} & \textbf{T} & \textbf{F} & \textbf{T} & \textbf{F} & \textbf{T} & \textbf{B} \\
\hline
U & Delay & \textbf{T} & \textbf{T} & \textbf{F} & \textbf{F} & \textbf{T} & \textbf{T} & \textbf{F} & \textbf{F} & \textbf{B} \\
    & Accept & \textbf{F} & \textbf{F} & \textbf{T} & \textbf{T} & \textbf{F} & \textbf{F} & \textbf{T} & \textbf{T} & \textbf{B}\\
    & Abandon & \textbf{F} & \textbf{F} & \textbf{F} & \textbf{F} & \textbf{F} & \textbf{F} & \textbf{F} & \textbf{F} & \textbf{T}\\
 \hline
L & Insist & \textbf{F} & \textbf{F} & \textbf{F} & \textbf{F} & \textbf{T} & \textbf{T} & \textbf{T} & \textbf{T} & \textbf{B} \\   
\hline
\end{tabular}
\end{table} 

\subsection{Interstate Conflicts-Russia Ukraine}
\begin{figure}[b]
      \centering
 \includegraphics[scale=0.65]{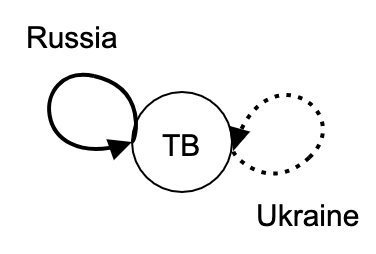}
  \caption{Graph Model of Russia Ukraine Conflict }
      \label{fig:RU}  
 \end{figure}

Interstate conflicts are generally analyzed in the $2 \times 2$ cooperate-defect game structure, in which equilibrium solutions obtained by the attitude of DMs are interpreted to have more escalatory or de-escalatory preferences.
In the case of a conflict with two DMs and one option for each (Y/N), even if the structure of the conflict originally has no equilibrium, some equilibrium solutions can be obtained by introducing a framework, for example, setting the permissible ranges \cite{kato-1} for the DMs' preference order. It is possible to consider a path toward consensus building using the resolution result as a clue.
General security discussions, such as the deterrence theory, are also based on these concepts.
However, we observed that the premise of such discussions might have been overturned by the recently erupted Russia--Ukraine conflict.
The states in \textsf{B4} and the actual status of the graph of the conflict are provided in Table \ref{tab:ru} and Figure \ref{fig:RU}, respectively. Assume that both DMs have a positive attitude toward self.

\begin{table}[b]
\caption{States and Options in Russia Ukraine Conflict}
\label{tab:ru}
\centering
\small
\begin{tabular}{c|c|c}
\hline
  {}    &  Russia&  Ukraine\\
\hline                           	
Option & Attack & Attack \\
\hline
$s_1$ & \textbf{T} & \textbf{B}\\

\hline
\end{tabular}
\end{table}

Sixteen states generated by the four-value combination of $2 \times 2$ games, as shown in Table \ref{tab:PD states}; 
Only four states generated from the Y/N (T/F) selection are subject to analysis in the general analysis method, whereas Russia: \textbf{T} and Ukraine: \textbf{B}, which may better reflect the current situation, are not considered.
If we had analyzed the conflict using a four-valued system that included \textbf{B} and \textbf{N}, we could predict a stalemate in \textbf{T} and Ukraine: \textbf{B}.
Within a system in which the nuclear weaponry fulfills a deterrent function, it could be possible to respond to some extent through a binary rational analysis; it can be said that the Russia--Ukraine conflict has already shown that such an approach is no longer viable.
Considering the current situation in which artificial intelligence is being introduced in the security field and for weapon control, this case suggests the danger of building a program with only assumptions on reasonable conditions.

\section{Conclusion}
\label{concl.}
In GMCR stability analysis, it is not a prerequisite that the option selection structure underlying the state formulation be precise because it is sufficient to know the DMs' preference orders concerning the states.
Nonetheless, introducing a paraconsistent logic system to the state definition may improve the computability and extensibility of GMCR analysis.
In particular, as demonstrated in Section \ref{sec:4}, it is likely to be extremely useful in decision-making systems in international relations and security. Four-valued logic, which encompasses contradictions, can be a more conservative system; there are many areas where its introduction would be worthwhile for analyzing real-world problems.

Future work may include reexamining the $2 \times 2$ international dispute structure by introducing the four-valued logic concept and generalizing it in the system of GMCR analysis based on studies of its relativeness to other extended concepts such as coalition \cite{coalition-1},\cite{coalition-2},\cite{coalition-3} permissible range \cite{kato-1}, \cite{kato-2}, and new reachability by external factors \cite{kato-3}.

\end{document}